\documentclass[verbose=true,letterpaper,a4paper,margin=2cm]{article}
\PassOptionsToPackage{margin=2cm}{geometry}

\usepackage{PRIMEarxiv}

\usepackage[utf8]{inputenc} 
\usepackage[T1]{fontenc}    
\usepackage{hyperref}       
\usepackage{url}            
\usepackage{booktabs}       
\usepackage{amsfonts}       
\usepackage{nicefrac}       
\usepackage{microtype}      
\usepackage{lipsum}
\usepackage{fancyhdr}       
\usepackage{graphicx}       
\usepackage{amssymb}        
\usepackage{authblk}
\usepackage{amsmath}
\usepackage{booktabs}
\usepackage{longtable}
\usepackage[table]{xcolor}
\usepackage[nameinlink]{cleveref}

\usepackage[style=numeric,backend=bibtex,maxnames=3,minnames=1]{biblatex}
\usepackage{makecell}
\usepackage{tikz}
\usepackage{geometry}
\usepackage{hyperref}  
\definecolor{darkblue}{rgb}{0.0, 0.0, 0.55}  
\usepackage[skip=0pt]{caption}  
\usepackage{float}
\usepackage{adjustbox}

\definecolor{rowgray}{gray}{0.95}  

\graphicspath{{media/}}     

\title{
  \begin{tikzpicture}[scale=0.8]
    \node[rotate=0, text=darkblue] at (0, 0) {\Huge\bfseries\ttfamily HiBayES}; 
    \node[rotate=0, text=black] at (0, -1.25) {\LARGE A Hierarchical Bayesian Modeling Framework};
    \node[rotate=0, text=black] at (0, -2.25) {\LARGE for AI Evaluation Statistics};
  \end{tikzpicture}
}

\author{
  \large\textbf{Lennart Luettgau}, \textbf{Harry Coppock}, \textbf{Magda Dubois},\\
  \large\textbf{Christopher Summerfield}, \textbf{Cozmin Ududec}
}
\affil{\normalsize UK AI Security Institute, London, UK}
\affil{\normalsize \texttt{\{firstname.lastname\}@dsit.gov.uk}}

\begin{document}
\maketitle

\begin{abstract} 
As Large Language Models (LLMs) and other AI systems evolve, robustly estimating their capabilities from inherently stochastic outputs while systematically quantifying uncertainty in these estimates becomes increasingly important. Further, advanced AI evaluations often have a nested hierarchical structure, exhibit high levels of complexity, and come with high costs in testing the most advanced AI systems. To address these challenges, we introduce \texttt{HiBayES}, a generalizable Hierarchical Bayesian modeling framework for AI Evaluation Statistics. \texttt{HiBayES} supports robust inferences in classical question-answer benchmarks and advanced agentic evaluations, particularly in low-data scenarios (e.g., $<$ 20 data points per evaluation). Built on Generalized Linear Models (GLMs), Bayesian data analysis, and formal model comparison, \texttt{HiBayES} provides principled uncertainty quantification and robust parameter estimation. This paper offers a comprehensive introduction to \texttt{HiBayES}, including illustrative examples, comparisons to conventional statistical methods, and practical guidance for implementing multilevel Bayesian GLMs. Additionally, we provide a \texttt{HiBayES} software package \cite{UKGovernmentBEIS2025hibayes} (Beta version) for out-of-the-box implementation. \end{abstract}

\section{Introduction}
As the capabilities of AI models (including Large Language Models, or LLMs, and agentic systems) improve \cite{kwa2025measuringaiabilitycomplete}, researchers face the increasingly important challenge of making statistically sound, reliable, and nuanced claims about model capabilities based on the results of evaluations \cite{Hendrycks2020,Liang2022,Chen2021,Shi2022,Yang2023,Yu2024}. Accurate assessment of model capabilities from evaluations, and the related prediction of their behavior in real-world use cases \cite{Tamkin2024} or human uplift \cite{Anthropic2024,OpenAI2024}, are not only crucial for driving further model development, but also to improve model safety and safeguards in deployment. However, the toolkit of conventional approaches to evaluation statistics currently available to AI researchers and practitioners severely limits their ability to draw robust conclusions, appropriately quantify uncertainty, and account for the complex hierarchical structure inherent in AI evaluation data.

Three common evaluation scenarios illustrate the limitations of current approaches:

\textbf{Scenario 1: Estimating overall performance.}
Researchers often need to evaluate the overall performance of a model, which is typically a composite of multiple measures. This includes claims such as "Model A has attained human-level performance on reasoning tasks" or "Model B has surpassed the 80\% safety threshold necessary for deployment". Conventional evaluation methods commonly report aggregated metrics, like 82\% overall accuracy, neither considering differences in the number of samples that were drawn for specific measures that contribute to overall performance nor quantifying uncertainty surrounding these estimates in a principled way. This practice leaves unanswered whether a model has truly achieved the threshold or if the observed performance lies within the margin of error of the threshold \cite{Anthropic2024,OpenAI2024}.

\textbf{Scenario 2: Model comparison across domains.} When comparing models across benchmarks covering distinct domains (e.g., reasoning, coding, cybersecurity), evaluations typically produce conclusions like "Model A outperforms Model B on reasoning tasks, while Model B excels at coding tasks". Such analyses often treat each domain independently, failing to account for the hierarchical structure of evaluation data and the correlations within (and between) domains, leading to overstated claims about domain-specific advantages of one model over another.

\textbf{Scenario 3: Capability elicitation in agentic benchmarks.} When assessing model capabilities on tasks measuring, for example, reasoning or software engineering, performance often varies dramatically with task difficulty and the effort invested in reasoning and scaffold development (e.g., chain-of-thought steps, multiple attempts, tool improvements). Researchers might claim "Model A performs better when using more reasoning steps" based on limited samples across varying difficulty levels \cite{Mialon2023}. Current approaches typically aggregate results without accounting for how task difficulty interacts with reasoning effort, and fail to properly model the hierarchical structure where responses are nested within difficulty levels. In these complex evaluation settings and low data regimes, conventional statistical approaches become unreliable for making strong claims about capabilities.

Current approaches predominantly follow the "highest-number-is-best approach" \cite{Miller2024}, where a model with the highest average performance metrics is considered the best model on a given benchmark, without rigorous statistical validation. These practices do not systematically quantify uncertainty, nor do they employ formal model comparisons, or account for the hierarchical nature of evaluation data -- where responses are nested within items/questions, which are nested within subdomains (e.g., MBPP \cite{Yu2024}, DS-1000 \cite{Lai2022}, BoolQ \cite{Clark2019}, and RACE-H \cite{Lai2017}, which in turn cluster items into domains (see \Cref{fig:hierarchical_structure}). Statistical models that do not acknowledge similarity between data clusters (e.g., evaluation domains or subdomains) and therefore do not share (partially pool) estimates across levels of the data hierarchy are known as "flat models". Such approaches discard information on the hierarchical data structure and have been shown to overestimate true effect size, resulting in overconfident conclusions unlikely to generalize to future data \cite{McElreath2020a,McElreath2020b}.

Evaluation data typically also contain different sources of variation (i.e., within- and between-LLM variation and between-task variation) at different levels of the data hierarchy. Additional sources of variation may extend to comparing different LLM families (e.g., different developers or systems – we might expect LLMs with similar parameterizations to perform similarly), comparing prompts (e.g., providing context may lead to consistently higher scores, Chain-of-Thought prompting may produce higher scores on average, etc.), and question clusters (some question may be more similar to each other than others, e.g., about the same topic). Additionally, it is possible to explicitly estimate the effects of testing conditions, or elicitation of LLM capabilities (e.g., sandbagging vs no sandbagging, comparing different agent scaffolds \cite{swebench2023}) or compare performance across repeated evaluation runs separated in time. Conventional statistical approaches (e.g., \textit{t}-tests) struggle with explicitly separating these sources of variation and quantifying where the uncertainty in the inference about LLM capabilities comes from, forcing evaluators to make decisions about which variables of interest to focus on, while aggregating over other variables (that could be of interest as well).

To overcome these issues, we introduce \texttt{HiBayES} -- a flexible, principled and generalizable statistical framework for AI evaluations, based on hierarchical (multilevel) Bayesian Generalized Linear Models, Bayesian uncertainty quantification and formal GLM comparison using information criteria. We take inspiration from modern approaches to statistics and statistical best practices in the natural and
social sciences \cite{McElreath2020a,McElreath2020b} to directly address the methodological shortcomings of conventional approaches by providing three key advantages: (1) principled uncertainty quantification, (2) explicit modeling of the hierarchical structure of evaluation data, and (3) robust inference for agent-based systems even in low data regimes, where conventional methods typically fail.

The introduction of the \texttt{HiBayES} framework is particularly timely given the economic constraints that increasingly limit evaluation data collection for new agent-based systems or reasoning models -- solving a single task in some benchmarks may easily amount to hundreds of US Dollars worth of tokens \cite{arcprize2025analyzing}. At the same time, agentic model capabilities and their evaluation substantially raise the stakes for accurate capability assessment \cite{metr-s-gpt-4-5-pre-deployment-evaluations}. By incorporating hierarchical modeling, appropriate treatment of data distributions, and formal model comparison through information criteria, \texttt{HiBayES} enables more reliable, nuanced, and scientifically sound evaluation of AI capabilities.

\begin{figure}[H]
    \centering
    \adjustbox{trim={0 6cm 0 0}, clip}{\includegraphics[width=1\textwidth]{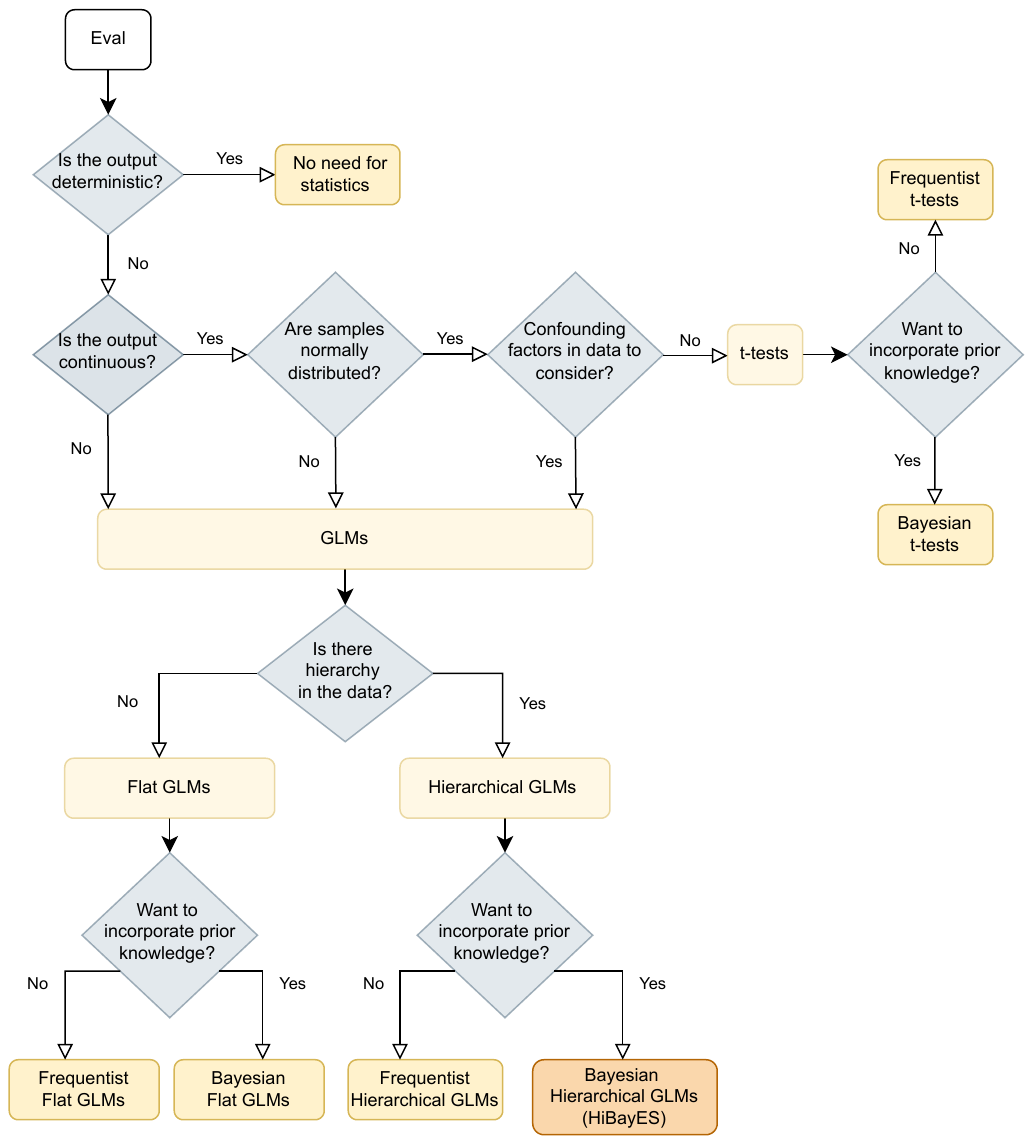}}
    \caption{Flowchart to determine which statistical approach to choose. More information on the differences between \textit{t}-tests and GLMs can be found in \Cref{tab:table1} and in \cite{McElreath2020a,McElreath2020b}.
}
    \label{fig:flowchart}
\end{figure}

After an introduction of the foundational concepts of \texttt{HiBayES}, we provide illustrative examples for the implementation of multilevel Bayesian GLMs and GLM comparisons. An extensive comparison of \texttt{HiBayES} and conventional evaluation statistics is presented in \Cref{tab:table1}, and a decision framework for selecting the appropriate statistical modeling approach is presented in \Cref{fig:flowchart}.
In \Cref{sec:appendix} (Appendix), we detail the steps for a workflow with \texttt{HiBayES}, guiding the reader to specify and fit their own multilevel Bayesian GLMs.

\begin{longtable}{p{2.5cm} p{6.25cm} p{6.25cm}}
\caption{Comparison of \texttt{HiBayES} and conventional evaluation statistics.} \\
\toprule
\textbf{} & \textbf{Multilevel GLMs (\texttt{HiBayES})} & \textbf{Conventional Approach (e.g., \textit{t}-tests)} \\
\midrule
\endfirsthead

\toprule
\textbf{} & \textbf{Multilevel GLMs (\texttt{HiBayES})} & \textbf{Conventional Approach (e.g., \textit{t}-tests)} \\
\midrule
\endhead

\rowcolor{rowgray}
\small \textbf{Dataset size} & \small Handles any dataset size \vspace{0.5em}\newline%
Stable estimates even in low data regimes with nested data structures due to partial pooling of estimates & 
\small Rule of thumb: $\geq$ 20-30 data points per data cluster/group
\vspace{0.5em}\newline%
Dependent-samples/pairwise comparisons: same number of data points per data cluster/group (need to pad smaller data vector with additional values or subsample at random from larger data vector)
\vspace{0.5em}\newline%
Strictly only valid with large quantities of data (stability of estimates based on asymptotic assumptions derived from central limit theorem)
\\

 \small \textbf{Assumptions about the data} & \small Data generated by distribution in exponential family (e.g., Normal, Binomial, Poisson)
\vspace{0.5em}\newline%
Hyperparameters of data distribution depend on a linear combination of independent variables (linear model) & \small Assumes approximately Gaussian data
\vspace{0.5em}\newline%
Real, continuous values range $[-\infty; +\infty]$
\vspace{0.5em}\newline%
Homogeneity of variance in data clusters/groups \\
 
\rowcolor{rowgray}
\small \textbf{Pre-averaging} & \small No data pre-averaging needed, works on obtained raw item-level scores
\vspace{0.5em}\newline%
Preserve natural variation across levels of the nested data hierarchy
\vspace{0.5em}\newline%
Partial pooling of parameter estimates across data clusters and levels
 & \small Data pre-averaging across repeats, evaluations or subdomains needed
\vspace{0.5em}\newline%
Degree of freedom: Requires decision on measure of central tendency (mean, median or mode) used for averaging
\vspace{0.5em}\newline%
Removes variance across repeats (and lower levels of the data hierarchy), prevents optimal propagation of variance across levels
\vspace{0.5em}\newline%
Increased likelihood of obtaining overconfident, biased estimates that generalize poorly
 \\
 
\raggedright \small \textbf{Repeated sampling and imbalances} & \small Parameter estimates adjusted for repeated sampling (e.g., if there is more than one observation from the same LLM) 
\vspace{0.5em}\newline%
Parameter estimates adjusted for sampling imbalances (e.g., some evaluations, domains etc sampled more often than others)
\vspace{0.5em}\newline%
(Partial) pooling of information across data clusters and levels enhances precisions of parameter estimation
& \small Not explicitly accounted for, no principled approach for correcting for such differences, may lead to varying precision in parameter estimation across data clusters/groups (pre-averaged data for evaluations or domains sampled more frequently will be more precise than for less sampled evaluations or domains)
\vspace{0.5em}\newline%
Problematic for assumption of homogeneity of variance in data clusters/groups
 \\
 
\rowcolor{rowgray}
\small \textbf{Flexibility} & \small Can capture hierarchical data and data structures of (nearly) arbitrary complexity & \small Limited to comparison of two data levels \\
 
\small \textbf{Prior specification} & \small Specification of prior distribution and choice of parameters required
\vspace{0.5em}\newline%
Degree of freedom, e.g., which distribution and which parameterization to use, informative vs non-informative priors
\vspace{0.5em}\newline%
Allows to explicitly and clearly express assumptions and “state of ignorance” (continuum of more or less plausible, logically consistent specifications)
 & \small Not required 
\vspace{0.5em}\newline%
(but: Maximum-likelihood estimation relies on hard-coded assumption of uniform distribution over posterior parameter values)
 \\
 
\rowcolor{rowgray}
\raggedright \small \textbf{Uncertainty quantification} & \small Principled and generalizable approach to uncertainty quantification across nested hierarchical levels of the data: Dispersion of posterior parameter distribution (posterior variance and Credible Interval (CI)/ Highest Posterior Density Interval (HPDI)) & \small Uncertainty of parameter estimates (point estimates) not quantified
\vspace{0.5em}\newline%
Requires (adjusted) variance estimators (e.g., clustered cluster-adjusted standard errors) to quantify uncertainty in data (no principled framework to capture uncertainty across nested hierarchical levels of the data simultaneously)
 \\
 
\small \textbf{Inference} & \small Supports inferences about differences of parameter estimates (if no overlap of CI/HPDI) as well as equivalence (if full overlap of CI/HPDI) & \small Null Hypothesis Significance Testing (NHST) only (what is the probability of obtaining the observed test static or more extreme values if the null hypothesis is true?): 
\vspace{0.5em}\newline%
Degree of freedom: need to define significance level $\alpha$ (e.g., $\alpha$ = .05)
\vspace{0.5em}\newline%
Inferences about differences possible (if $p < \alpha$), but results inconclusive/unclear if $p \geq \alpha$ \\

\rowcolor{rowgray}
\raggedright \small \textbf{Multiple comparisons correction} & \small No adjustment of estimates needed, all parameters are estimated simultaneously using a single GLM & \small No straightforward generalization to scenarios involving more than two levels (e.g., LLMs/evaluations/domains/subdomains)
\vspace{0.5em}\newline%
Need to adjust \textit{p}-values for any scenario involving running multiple \textit{t}-tests (e.g., using Family-wise Error or False-Discovery Rate Correction) \\
 
\small \textbf{Model comparison} & \small GLM predictive accuracy quantified using log-likelihood and information criteria (e.g., WAIC), statistical models can be compared to alternative models 
\vspace{0.5em}\newline%
Supports generalization of parameters and inferences to future data
& \small Not provided, no generative model of the data, no statistical model comparison possible (lacks predictive accuracy)
\vspace{0.5em}\newline%
No generalization to future data possible
 \\
 
\rowcolor{rowgray}
\small \textbf{Usage difficulty} & 
\small Implementation: - medium (need to specify GLM for respective use case, MCMC sampling using packages like NumPyro)
\vspace{0.5em}\newline%
Quality control: - high (need to go through a series of controls and checks to ascertain GLM convergence and validity of inference)
\vspace{0.5em}\newline%
Interpretation: - medium (learn how to interpret mean and variance of posterior distribution and how to use CI/HPDI for making inferences; advantage: one quantity for all inferences)
 &
\small Implementation: - low (one-liner, but need to manually specify clusters and pre-average data)
\vspace{0.5em}\newline%
Quality control: - high (will always converge, but no indicators of fit and appropriateness, difficult to ascertain test validity or violation of assumptions)
\vspace{0.5em}\newline%
Interpretation: - medium (learn how to interpret \textit{t}- and \textit{p}-values, how to adjust for multiple comparisons, need to run a set of analyses if interested in more than one inference)
\\
\bottomrule
\label{tab:table1}
\end{longtable}

\section{Introducing \texttt{HiBayES} – A Hierarchical Bayesian Modeling Framework for AI Evaluation Statistics}
\subsection{Multilevel (hierarchical) linear modeling using Generalized Linear Models}

Generalized Linear Models (GLMs \cite{Nelder1972}) are highly flexible statistical models used to analyze the relationships between outcome variables and one or more predictor variables when the outcome variable distribution is not necessarily Gaussian. They can represent various types of data such as binary, count, or continuous outcomes. 

GLMs extend linear regression by allowing the outcome variable to follow different distributions. They model a linear combination of input features, which is then transformed by a \textit{link function} to match the scale of the outcome variable. The link function ensures that the model's predictions align with the statistical properties of the chosen distribution. The outcome variable is assumed to be generated by a probability distribution in the exponential family (e.g., Normal, Binomial, or Poisson). The parameters of the outcome data distribution depend on a linear combination of independent variables (linear model). 

An example GLM with a Binomial likelihood function and linear model based on a linear combination of predictors $X_1$ to $X_k$ and parameters $\beta_k$ is provided in multiple regression (distributional) notation in \Cref{eq:glm}.

\begin{equation}
\begin{split}
    Y_i &\sim \text{Binomial}(1, p_i) \\
    \text{logit}(p_i) &= \beta_0 + \beta_1 X_{1i} + \cdots + \beta_k X_{ki}
\end{split}
\label{eq:glm}
\end{equation}

\begin{flushleft}
\small where $Y_i$ denotes the number of successes in a binomial distribution. In the specific case where $n = 1$, $Y_i$ corresponds to a Bernoulli trial, resulting in a binary outcome. The success probability $p_i$ is linked to the linear predictor via the logit link function.
\end{flushleft}

\begin{figure}[htbp]
    \centering
    \includegraphics[width=1.025\textwidth]{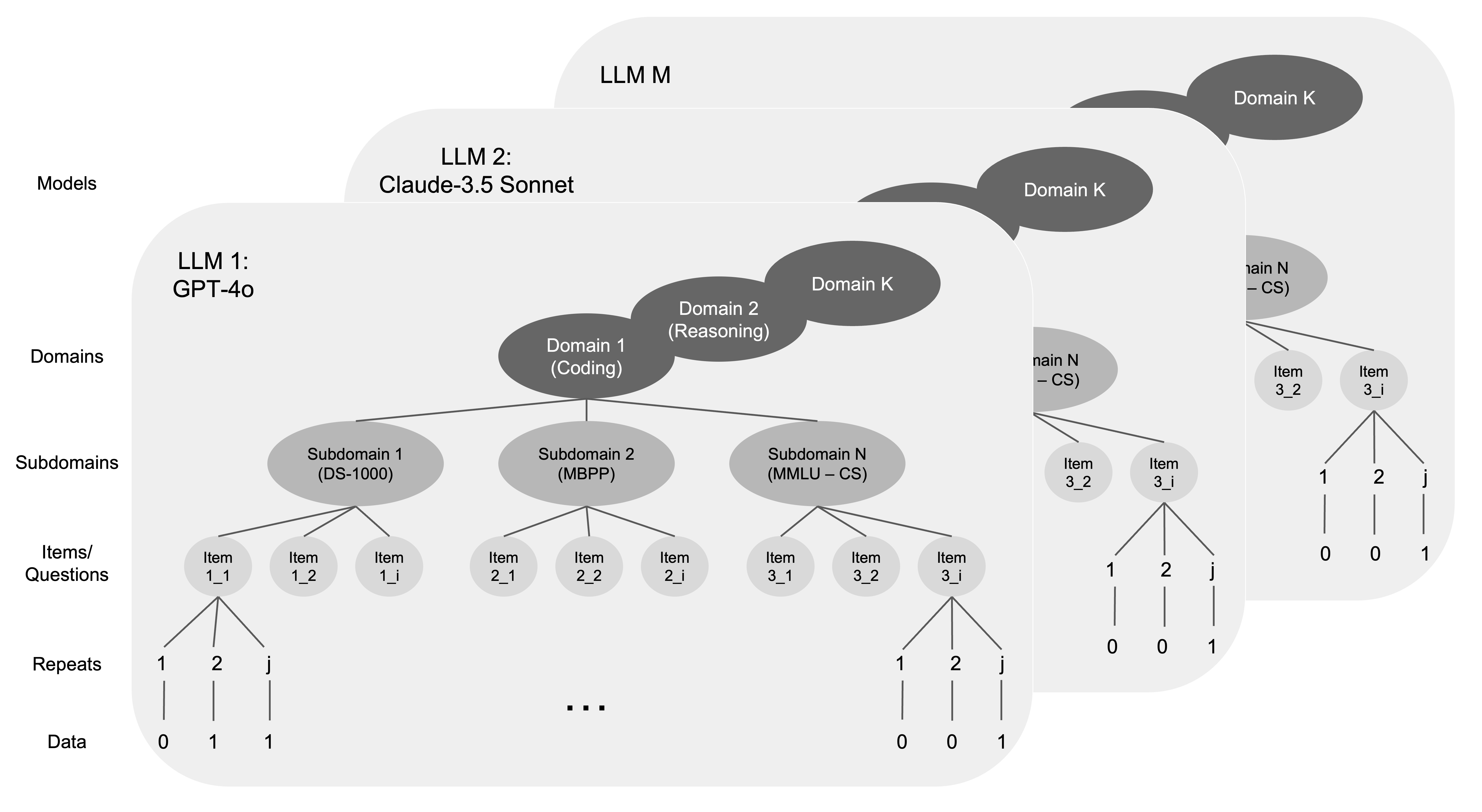}
    \caption{\textbf{Example hierarchically nested structure of evaluation data.} Data in evaluation scenarios are typically acquired across different, hierarchically nested levels. These levels comprise multiple repeats (1...j), of items (1...i), of different subdomains (1...N), which result from different domains (1...K). Typically, several LLMs (1...M) are evaluated on the same set of tasks. This hierarchical data structure is often not accounted for in current evaluation statistical practices. 
}
    \label{fig:hierarchical_structure}
\end{figure}

\Cref{eq:glm} represents a generic example of GLMs with a Binomial likelihood function, that does not explicitly account for a specific data acquisition context. However, evaluations are typically performed across different LLMs (e.g., comparing a SOTA LLM to one or more other LLMs), domains, subdomains, each comprising several items (questions/prompts), repeats (repeated attempts on a given domain/subdomain/item, see \Cref{fig:hierarchical_structure})

In practice, an important initial step is understanding the distribution of the outcome variable (data) and what kind of correlation structure is of interest to answer a research question. Typical evaluation scenarios (e.g., MMLU \cite{Hendrycks2020}, MGSM \cite{Shi2022} or InterCode CTF \cite{Yang2023}) involve responses to a given item that can either be correct or incorrect (e.g., pass/fail tasks with 1/0 outcomes, i.e., a series of independent Bernoulli trials). An adequate model of the data generating process of such binary outcome evaluations needs to explicitly model (a series of independent) Bernoulli trials – i.e., by using a Binomial likelihood function to estimate the success of a model in a given attempt. This approach ensures that the obtained parameter estimates generalize to future evaluations of the same LLM. 

GLMs can be extended to capture hierarchically nested data structures, through multilevel/hierarchical linear models (MLM). MLMs were developed to estimate complex, nested hierarchically structured single response data – mirroring exactly the data acquisition context of typical evaluation scenarios. MLMs straightforwardly allow parameters to be nested across levels of the data hierarchy, i.e., to simultaneously estimate the success rate of different LLMs across evaluations, domains, subdomains, repeats, and to quantify the uncertainty around these estimates. MLMs excel in evaluations with varying available data quantities and maintain stable and valid parameter estimates in low data regimes (e.g., $<$20 data points), in situations where data are grouped or clustered, and where these clusters differ from each other across various levels \cite{McElreath2020a,McElreath2020b}. MLMs also allow to separate the above sources of variation, to (partially) pool information across data clusters, and compute errors/uncertainty for each cluster in the data hierarchy. This in turn enhances the precision of parameter estimation, allowing us to learn more from the data than with flat GLMs (e.g., a low score on one of the reasoning benchmarks will inform the score computed on another reasoning benchmark).

\subsection{Bayesian data analysis for uncertainty quantification}

The majority of current evaluation reports do not present metrics of uncertainty, but focus on reporting point estimates (central tendencies of distributions, e.g., means, or statistical estimates, e.g., \textit{t}-values). Such an approach does not straightforwardly allow for rigorous quantification of uncertainty of the statistical quantities and does not support formalizing domain knowledge or prior experience to make rigorous statistical decisions. This lack of comprehensive uncertainty quantification can have severe consequences in under- or overestimating LLM capabilities both in testing environments and in production. This in turn may lead to overconfident or misleading interpretations of the capabilities of a LLM. AI Safety frameworks created by LLM developers, like the AI Safety Levels (ASL), introduced by \href{https://www.anthropic.com/news/anthropics-responsible-scaling-policy}{Anthropic’s Responsible Scaling Policy} focus on definitions of point values at which critical safety thresholds are surpassed. While such an approach promises objectivity and rigorous guidance in decision making, threshold-based frameworks in combination with statistical point estimates struggle to define principled decision criteria. 

To illustrate this point, let’s assume we specified a threshold of 25\% correct responses on a given benchmark as a key metric at which we should be concerned about a given LLM’s capabilities. Aside from the challenge of choosing a (potentially arbitrary) threshold value in the bounds [0; 100], how and when do we decide that this threshold has been crossed? One might say that the threshold is crossed strictly at any performance value >25\% correct responses. If a LLM achieves 75\% correct responses, few people would argue that such levels of performance are concerning, relative to the defined threshold. However, it is much less clear how to decide at near-threshold values. Would we consider a performance level of 24.9\% to be entirely harmless, whereas 25.1\% as risky? Most people will agree that even minor deviations from the threshold could be equally concerning if they happen below or above the threshold value. Point estimates provide no guidance on how to decide in such “close call” scenarios, as they do not consider the uncertainty of the estimated statistical quantity. We argue that probabilistic judgment and decision criteria are needed to tackle these questions in a principled manner. 

Recent work \cite{Miller2024} acknowledges the above intuitions and recommends the use of descriptive (e.g., error bars for uncertainty quantification) and inferential statistics for evaluations (e.g., \textit{t}-tests and confidence interval comparisons for statistical inferences). While we agree that adding error bars to point estimates and performing \textit{t}-tests to compare two LLMs’ performance levels on a given evaluation is a starting point for improving statistical practices, we caution that even adopting such practices will not fix the limitations of the conventional approach to evaluation statistics. This is because the suggested methods rely on a number of assumptions about the data and the data-generating process that are not met in typical evaluation scenarios (e.g., asymptotic assumptions, normality (of data/residuals), homogeneity of variance, etc.). Among these limitations, the most challenging one stems from the large amounts of data needed to satisfy asymptotic assumptions derived from the central limit theorem underlying the proposed approach for uncertainty quantification and the use of \textit{t}-tests. These methods rely on assumptions of Normality of the data (i.e., real and continuous in range $[-\infty; +\infty]$, drawn \textit{i.i.d.} from a Gaussian distribution; approximately 20-30 data points per evaluation/domain/subdomain) and – in the case of \textit{t}-tests – that data clusters/groups to be compared exhibit homogeneity of variance. Further, the increasing costs for generating data points in evaluating new agent-based AI systems will increasingly lead to situations where evaluation data is (due to economic constraints) extremely scarce ($<$ 20-30 data points per evaluation). Relying on the proposed methods for uncertainty quantification and statistical inference in low-data regimes will likely result in imprecise estimation of statistical quantities, leading to misguided data-driven inferences on LLM capabilities.

The proposed hierarchical Bayesian modeling framework naturally quantifies uncertainty across levels of the data hierarchy by using intervals of defined mass from the posterior parameter distribution (Credible Interval (CI) or Highest Posterior Density Interval (HPDI)).

\subsection{Formal GLM comparison using information criteria}

Finally, it is often not sufficient to compare mean performance estimates of two or more LLMs on a given evaluation (as done by \textit{t}-tests or by comparing GLM parameter HPDIs). This is because there is not only uncertainty about the parameter estimates, but more importantly, there is uncertainty about which data-generating process is most adequate for explaining the observed data (i.e., does the GLM fit the data well?). Conventional statistical approaches do not support answering research questions regarding predictive accuracies of different statistical models that incorporate distinct hypotheses about the data-generating process. A fundamental objective of statistical modeling is to capture the underlying data-generating process to estimate a models' fit to the data and its ability to generalize to future data, beyond the currently tested items and scenarios. This is at the heart of making inferences derived from evaluation results: We want to predict the performance of AI systems in real-world deployment contexts. Additionally, we are often interested in understanding how different variables (predictors) influence the predictions of our statistical model. GLM comparison formally tests various hypothesized data-generating processes and correlation structures, allowing for insights that \textit{t}-tests and other non-generative modeling frameworks do not provide \cite{McElreath2020a,McElreath2020b} .

\section{Use cases}
\subsection{Practical considerations for fitting GLMs using \texttt{HiBayES}}
\begin{figure}[htbp]
    \centering
    \includegraphics[width=1\textwidth]{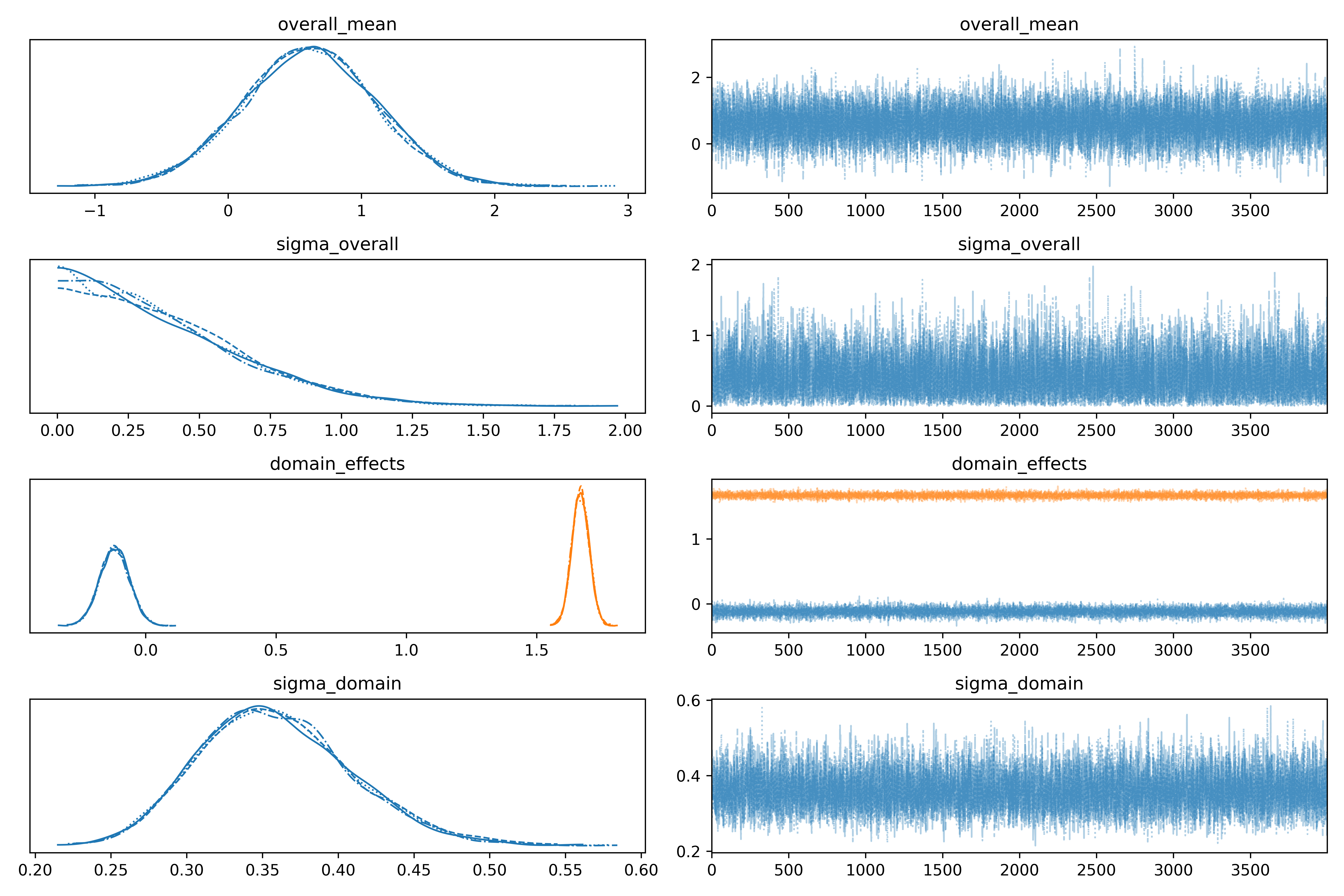}
    \caption{\textbf{Trace plots used to assess convergence of the MCMC chains for use case 1.} The left column shows the sampling distributions of the four Markov chains (on the logit scale). In the \texttt{domain\_effects} parameter, colors (blue/orange) indicate the two tested domains. The right column displays the trace plots for the estimated parameters, which show well-mixed MCMC chains – indicating convergence of the GLM, alongside other diagnostics.}
    \label{fig:trace_plots}
\end{figure}

To ensure reliable and robust inference, statistical modeling using \texttt{HiBayES} involves 5 aspects: 
\begin{enumerate}
    \item GLM specification (and prior predictive checks).
    
    \item Data preparation and parameter estimation using MCMC sampling.
    
    \item Assessment of GLM convergence and quality control. GLM convergence is indicated by the following metrics: 
        \begin{itemize}
            \item There should be very little (ideally 0) divergences.
            
            \item The Gelman–Rubin criterion ($\hat{R}$ or \texttt{r\_rhat}) should be close to 1 (ideally 1.00) for each sampled parameter.
            
            \item The number of effective samples (\texttt{n\_eff}) is a statistical estimate of how many independent samples the dependent samples are equivalent to after accounting for autocorrelation. This number gives an indication of how well-defined the GLM is and how smoothly the sampling process proceeded. This should be $\geq$ the number of MCMC samples drawn per chain (e.g., 4000). In general, a higher \texttt{n\_eff} is indicative of a well-specified and efficient model that produces efficient sequential samples.
            
            \item Trace plots should indicate noisy mixing of the MCMC chains – they should look like hairy caterpillars (e.g., see \Cref{fig:trace_plots})).
            
            \item Additionally, GLM convergence can be assessed using posterior predictive checks (not covered here).
        \end{itemize}

    \item Posterior sampling to quantify posterior mean estimates and uncertainty (intervals of defined mass, e.g., 95\% HPDI).

    \item Depending on the research question at hand, formal GLM comparison using information criteria.
\end{enumerate}

In \texttt{HiBayES}, we define GLMs and estimate the posterior distribution of the GLM parameters, using NumPyro \cite{Phan2019} for Markov Chain Monte Carlo sampling (using No-U-Turn-Sampler – NUTS – a variant of Hamiltonian Monte Carlo \cite{Hoffman2011}) (an alternative package for MCMC estimation is PyMC). 
As a reasonable default, we use 2000 warmup samples (which are used for optimization of the sampling process and not used for parameter estimation) and 2000-4000 inference samples (used for parameter estimation). To estimate reliable parameter distributions and to ensure GLM convergence, four separate Markov chains are selected and fit on four CPU cores in parallel.
After sampling is completed, we print a summary table for the parameter estimates and trace plots that provide us with diagnostic checks to perform quality control and ensure convergence of the inference. 

Estimation in \texttt{HiBayES} is done directly using samples from the posterior parameter distribution. After obtaining samples from the posterior distribution, several metrics can be used to provide insights on the quantities of interest. First, it is useful to investigate the LLM’s average performance (e.g., posterior mean). Second, the posterior parameter distribution provides a natural and principled quantification of the uncertainty of the estimates. This is commonly done by computing the standard deviation or variance of the posterior distribution or by computing an interval of defined mass, e.g., the (95\%) CI of the parameter estimate (range of values that contain the specified posterior mass), or the HPDI \cite{Gelman2013}), which is defined as the narrowest interval of parameter values that contains a specified proportion of the posterior density/mass (e.g., 95\%) \cite{McElreath2020a,McElreath2020b}. The latter metric gives an indication of which parameter values are most consistent with the observed data. Both CIs and HPDIs are useful to determine overlap of the estimate with given threshold values (e.g., whether performance is higher/lower than random guessing/chance level, or higher than 0/lower than 1), whether it is different from a defined threshold value (e.g., 25\%) or whether two LLMs perform differently on a given evaluation. 

We use information criteria, like the Widely Applicable Information Criterion (WAIC \cite{WatanabeSWATANAB2010}) to compare differently parameterized GLMs to estimate which GLM explains the data better. WAIC weights the log-likelihood of the GLM with the number of effective parameters to compute an estimate of the predictive accuracy on new data, equivalent to cross-validation \cite{WatanabeSWATANAB2010}.

\begin{figure}[htbp]
    \centering
    \includegraphics[width=.75\textwidth]{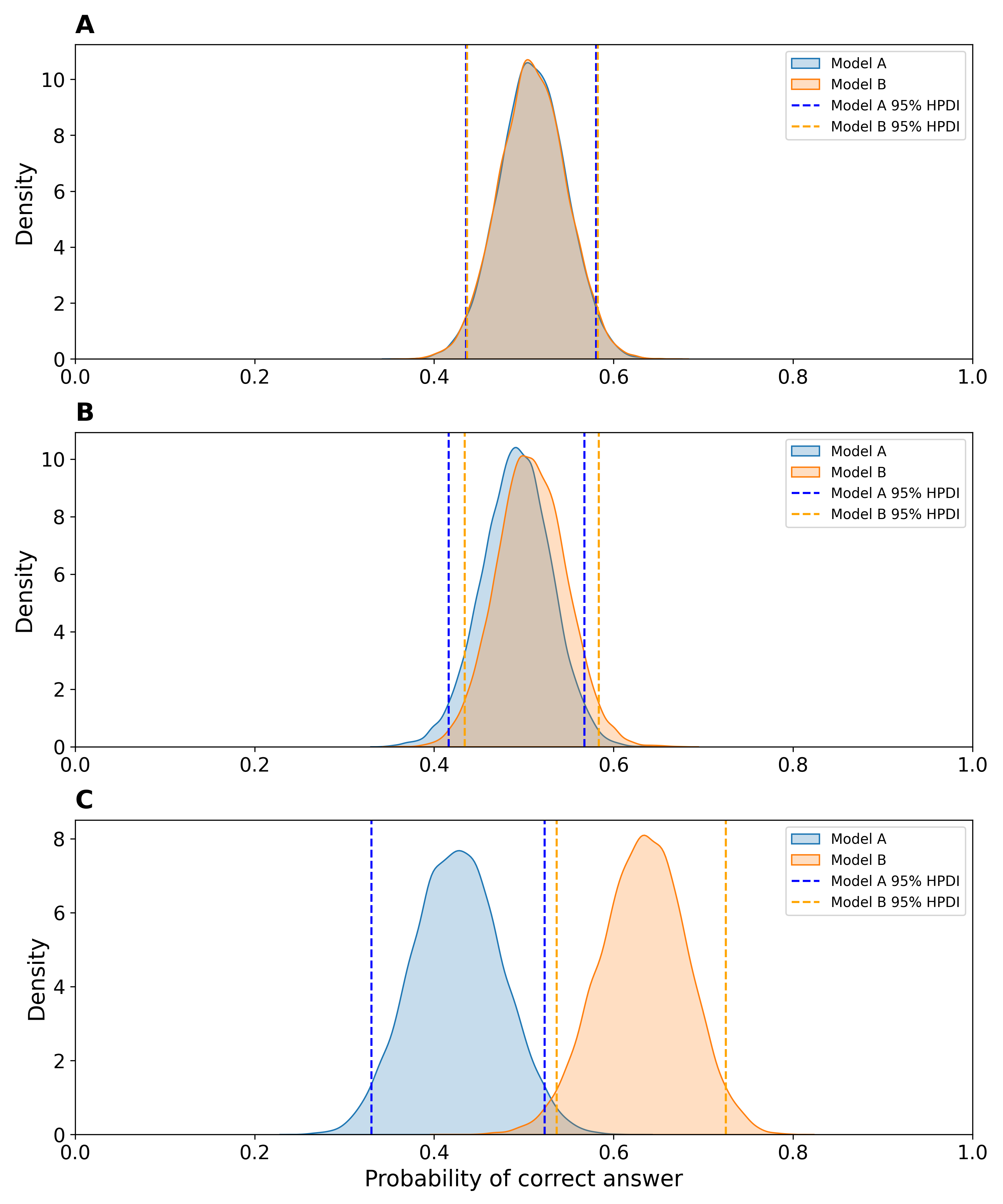}
    \caption{\textbf{Highest posterior density intervals (HPDIs) used for hypothesis testing.} Depicted are posterior probability distributions estimating two LLMs’ average performance and 95\% HPDIs of these posterior distributions (both on probability scale) from simulated data fitted with a GLM. The HPDIs represent the interval of parameter values that is most consistent with the data – the range of parameter values that the true performance score is most likely to fall into. HPDIs in panel A overlap completely (>99\% overlap), which suggests that LLM performance is equivalent (LLM A == LLM B). In panel B, HPDIs overlap partially ($\sim$88\% overlap), hence the result is inconclusive (LLM B <> LLM A). In panel C, HPDIs do not overlap (0\% overlap), so we can conclude that LLMs perform differently (LLM B > LLM A).
}
    \label{fig:hpdi_comparison}
\end{figure}

A great advantage of the Bayesian statistical modeling framework and the use of intervals of defined mass is that these metrics not only provide evidence for differences by rejecting a null hypothesis (as in Null Hypothesis Significance Testing (NHST) frameworks), but can also provide evidence for equivalence. For instance, we might want to know if two LLMs perform equivalently on a given evaluation (evidence for no difference). If the HPDIs around both LLMs’ performance estimates do not overlap, we can say the LLMs perform differently ( \Cref{fig:hpdi_comparison}C). If the HPDIs overlap partially, the result is inconclusive ( \Cref{fig:hpdi_comparison}B). However, if the HPDIs overlap completely (the narrower HPDI is fully contained in the wider HPDI, \Cref{fig:hpdi_comparison}A), we can say that LLM performance is equivalent \cite{Gelman2013}. In other words, these procedures provide support for the null hypothesis (that LLMs perform the same)! This is a great advantage over NHST, which precludes any conclusion in case \textit{p} $\geq$ .05 (or whatever your significance level $\alpha$ may be) – rendering statistical decision making in such situations unclear.

A common criticism of Bayesian methods is that the choice of the prior is subjective and that it is not straightforward to know if the choice is reasonable or valid. Indeed, Bayesian statistics require specifying a prior distribution and choosing parameters, creating researcher degrees of freedom (e.g., which distribution to choose, which parameterization to use, do we use informative or non-informative priors?). However, since these choices are explicitly stated, it is possible to clearly express assumptions and the "state of ignorance" before running the analyses, such that other researchers may criticize and scrutinize these choices. Additionally, the choice of the prior lies on a continuum between more and less plausible, logically consistent and inconsistent specifications. Based on past experience (e.g., with LLMs from the same family or developer, or other previous evaluations in similar domains) we have a sense for how a given LLM might perform on a given evaluation – this enables us to make a reasonable guess about the range of values that the posterior parameter distribution or its mean may fall into. We also know that certain parameter values are more or less likely based on the difficulty of the evaluation. For example, an easy evaluation task will probably result in a higher percentage of correct answers than a difficult evaluation task, so if we got the opposite result we would probably be very suspicious and would not trust the result. Why not explicitly state this prior knowledge? A uniform/non-informative prior – incorporating the assumption that all possible performance levels are equally likely for the given evaluation – can often not be upheld as a reasonable choice. To prevent overly restrictive assumptions, a common practice is to specify mildly informative priors for the parameters – preventing overconfident GLM specifications that are difficult to overturn by data. As an additional level of control and to assess how reasonable the choice of our prior is, we recommend running prior predictive checks to verify that the GLM-implied parameter pattern makes sense. 

\subsection{Datasets}
We prepared two LLM evaluation datasets using the UK AI Security Institute’s Inspect \cite{UKGovernmentBEIS2024inspectai} evaluation framework. 
Dataset 1 contains two LLMs (GPT-4o [release date: 01/06/2024] and Claude-3.5-Sonnet [release date: 20/06/2024]) performing four different benchmarks that are structured across two different domains, with each domain featuring two subdomains (see \Cref{fig:hierarchical_structure} for schematic):

\begin{enumerate}
  \item Coding (Subdomains 1A: MBPP \cite{Yu2024} (multiple repeats); 1B: DS-1000 \cite{Lai2022} ),
  \item Reasoning (Subdomains 2A: BoolQ \cite{Clark2019}; 2B: RACE-H \cite{Lai2017} ).
\end{enumerate}

Dataset 2 contains data from several LLMs (o1, o3-mini, Claude-3.7-Sonnet and GPT-4.5-preview) in an agentic evaluation (GAIA \cite{Mialon2023}). This benchmark features several tasks at varying levels of difficulty. The LLMs were run at different levels of reasoning effort. Due to token costs for higher levels of reasoning, this evaluation is performed in a low data regime (10 repeats per task). 

In the next section, we use the above datasets to demonstrate examples of common use cases in evaluation statistics.

\subsection{Use case 1: One LLM evaluated on multiple domains (benchmarks) with varying numbers of items}

One of the most common AI evaluation scenarios is estimating whether a model has achieved a capability threshold. Current approaches typically report that a model achieves a certain \% accuracy without quantifying uncertainty, leaving open the question of whether the model has truly crossed the threshold or whether the observed performance is within the margin of error of the threshold. Here, we focus on a use case to assess domain-specific LLM performance levels, as well as overall LLM performance, and quantify uncertainty around the performance estimates. We outline the performance of Claude 3.5 Sonnet on coding tasks (domain 1) and reasoning tasks (domain 2) -- without explicitly considering the nested subdomains. For data efficiency purposes and in order to consider performance for unique items, we only include response for the first repeat of each evaluation in our analyses. For domain 1, we recorded \textit{N} = 1257 responses for unique items, and for domain 2, \textit{N} = 6768 responses for unique items. Such item imbalances across evaluations are very common and complicate valid comparisons in conventional statistical approaches. 

To compare LLM performance on these evaluations, in a conventional approach researchers would typically conduct a dependent-samples \textit{t}-test (repeated-measures \textit{t}-test: the same LLM is tested repeatedly across different domains). In conventional statistics, we ask whether the \textit{t}-value for the domain difference is statistically significant ($p < 0.05$). However, different sample sizes across domains pose a serious challenge for the validity of dependent-samples \textit{t}-tests, as these require the two data vectors to have equal length. To perform the \textit{t}-test, we must either pad the smaller vector with additional values or subsample randomly from the larger one. Both solutions are suboptimal, as they require us to either inflate the data or discard valid evaluation data. In either case, we risk overestimating the true difference between the two domains.

Due to the large sample sizes in our dataset, standard errors of the mean (SEMs) are small and  empirical means are estimated with a high degree of certainty. The resulting \textit{t}-value is very large ($t = 27.30$, $p < .001$), suggesting a large difference between domain 1 and domain 2. Even smaller differences between domain-specific performance levels would likely have been deemed significant under such small SEMs. When working with large datasets, decision criteria based on \textit{p}-values from \textit{t}-tests are not sensitive enough to meaningfully reflect effect magnitudes.

To compare LLM performance using \texttt{HiBayES}, we specify a hierarchical GLM with partial pooling across domains (\Cref{eq:use_case1_eq})

\begin{equation}
\begin{aligned}
\text{Correct}_j &\sim \text{Binomial}(n_j, p_j) \\
\text{logit}(p_j) &= \text{domain}_j \\
\text{domain} &\sim \text{Normal}(\bar{\mu}_{\text{domain}}, \sigma_{\text{domain}}) \\
\bar{\mu}_{\text{domain}} &= \mu_{\text{overall}} + \sigma_{\text{overall}} \cdot z_{\text{overall}} \\
\mu_{\text{overall}} &\sim \text{Normal}(0, 1) \\
\sigma_{\text{overall}} &\sim \text{HalfNormal}(0.5) \\
z_{\text{overall}} &\sim \text{Normal}(0, 1) \\
\sigma_{\text{domain}} &\sim \text{HalfNormal}(0.1).
\end{aligned}
\label{eq:use_case1_eq}
\end{equation}

\begin{flushleft}
\small \textit{}$\text{Correct}_j$ denotes the number of correct answers for the $j$-th domain, and is distributed as the proportion of correct answers ($p_j$) among $n_j$ trials. The variable $\text{domain}_j$ models average performance for domain $j$. The GLM assumes domain-specific parameters are drawn from a domain-level normal distribution, with mean $\bar{\mu}_{\text{domain}}$ and standard deviation $\sigma_{\text{domain}}$. The mean $\bar{\mu}_{\text{domain}}$ is drawn from a reparameterized hyper-distribution governed by $\mu_{\text{overall}}$, $\sigma_{\text{overall}}$, and $z_{\text{overall}}$, all using mildly informative priors.
\end{flushleft}

Note that the GLM definition features a non-centered parameterization. This is a reparameterization technique that improves sampling efficiency and convergence by transforming latent variables into components of their mean and standard deviation, often using auxiliary variables. Specifically, $\bar{\mu}_{\text{domain}}$ is defined as a combination of $\mu_{\text{overall}}$ and the product of the standard deviation $\sigma_{\text{overall}}$ and the auxiliary variable $z_{\text{overall}}$ (which is sampled from a standard Gaussian distribution). This decomposition avoids inefficiencies in the sampling process (as indicated by slow sampling and a high number of divergences) that sometimes occur if the shape of the posterior distribution is complex (as is often the case when using varying effects or hierarchical GLMs). GLMs can be specified with a centered parameterization as well, but a non-centered parameterization is often beneficial when fitting more complex GLMs.

\begin{figure}[htbp]
    \centering
    \includegraphics[width=.85\textwidth]{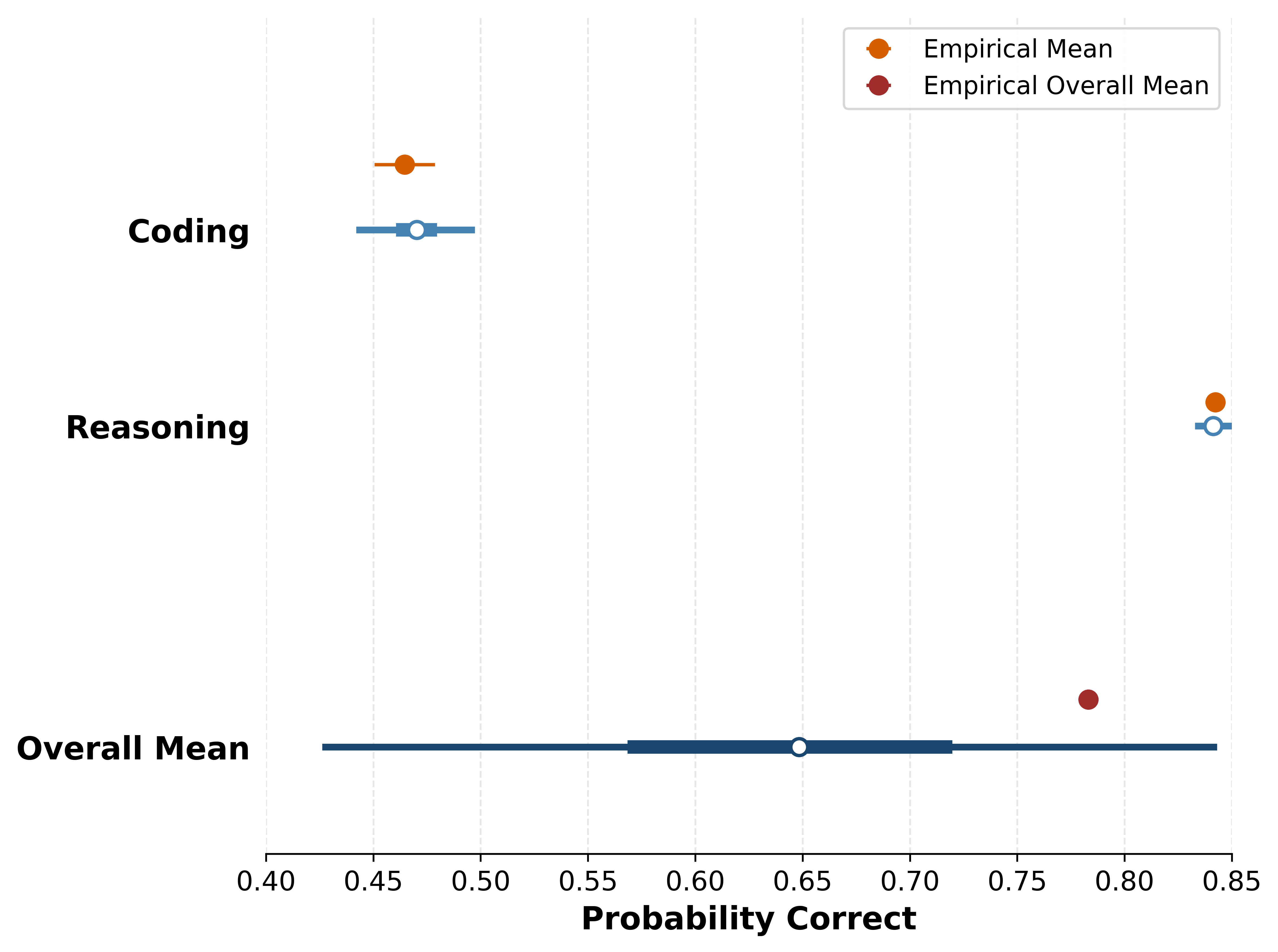}
    \caption{\textbf{Forest plot to display posterior means, posterior distributions, and 95\% HPDI (on probability scale) for use case 1.} Focusing on domain-specific and overall performance effects (blue), this plot indicates the uncertainty (vertical lines, 95\% HPDI), around the posterior mean (dots) for the respective parameter estimates. Additionally, for comparison between \texttt{HiBayES} and conventional evaluation statistics approaches, we present empirical means and SEM per domain (orange) and for overall (red).}
    \label{fig:use_case1_forestplot}
\end{figure}

We use forest plots to display the posterior distributions and compare GLM estimated domain effects and the uncertainty of the estimates (HPDI) and overall mean parameter to empirical means (arithmetic means) and standard errors of the mean (SEM) for each domain and overall performance (\Cref{fig:use_case1_forestplot}). 

While the \texttt{HiBayES} GLM estimated domain-specific effects and the empirical means per domain are similar, the empirical SEMs are much narrower than the GLM estimated HPDIs. This may be misleading and produce overly confident interpretations of LLM capabilities, as the empirical SEM provides too narrow estimates of uncertainty. However, an even larger risk of overstated capability interpretations stems from the overestimation of overall performance levels using the empirical mean. Here, the estimate is influenced strongly by the larger number of datapoints from domain 2 ($N = 6768$) vs domain 1 ($N = 1257$). While this could potentially be mitigated by averaging per domain first and then computing the overall performance, such an approach will not be reliable for estimating the SEM. Using \texttt{HiBayES} the overall performance is estimated exactly at the central point between Coding and Reasoning estimates (as one would intuitively expect) and the HPDI spans the width of both domain-specific HPDIs – the overall mean estimate regularizes the domain-specific estimates.

Finally, for inferences about which domain-specific performance is higher, in \texttt{HiBayES} we investigate the (non)overlap of the HPDIs of the posterior parameter distributions as a means to infer statistically meaningful differences. This approach generalizes to all settings. Here, performance in domain 2 (posterior mean = .84, 95\%-HPDI = $[.83; .85]$) is higher than in domain 1 (posterior mean = .47, 95\%-HPDI = $[.44; .49]$), since the HPDIs for domain 1 and domain 2 do not overlap.

Parameter estimates from \texttt{HiBayES} can be directly interpreted as effect sizes, and Bayesian data analysis ensures mathematically optimal handling of evidence (given the prior distribution). Consequently, statistical inference in \texttt{HiBayES} is far less sensitive to sample size imbalances than conventional statistical approaches.

\subsection{Use case 2: Compare two LLMs evaluated on multiple domains and subdomains with varying numbers of items.}

Another common AI evaluation scenario is comparing models across different evaluation benchmarks, investigating domain-, subdomain-specific, and overall performance level differences between two or more LLMs (e.g., new LLM vs current SOTA model) and quantifying uncertainty around performance estimates. In this illustrative example, we compare the performance of Claude 3.5 Sonnet and GPT-4o on domain 1: coding (subdomains 1A: MBPP and 1B: DS-1000) and domain 2: reasoning (subdomains 2A: BoolQ and 2B: RACE-H). For data efficiency purposes and in order to consider performance for unique items, we again only include responses for the first repeat of each evaluation in our analyses. For subdomain 1A and 1B, we recorded $N = 257$ and $N = 1000$, respectively, and for 2A and 2B, $N = 3270$ and $N = 3498$, respectively.

To compare LLM performance on these benchmarks, conventional statistics would typically use two-sample/independent-sample \textit{t}-tests. We find the conventional approach suggests GPT-4o performs better than Claude 3.5 Sonnet in DS-1000 ($t = 6.02$, $p < .001$) and BoolQ ($t = 5.18$, $p < .001$). 

We additionally specified a hierarchical GLM with partial pooling across domains (\Cref{eq:use_case2_eq}).

\begin{equation}
\begin{aligned}
\text{Correct}_{i,j,k} &\sim \text{Binomial}(n_{i,j,k}, p_{i,j,k}) \\
\text{logit}(p_{i,j,k}) &= \text{subdomain}_{i,j,k} \\
\text{model}_i &= \mu_{\text{overall}} + \sigma_{\text{model}} \cdot z_{\text{model}_{i}} \\
\mu_{\text{overall}} &\sim \text{Normal}(0, 1) \\
\sigma_{\text{model}} &\sim \text{HalfNormal}(0.1) \\
z_{\text{model}_{i}} &\sim \text{Normal}(0, 1) \\
\text{domain}_{i,j} &= \text{model}_{i} + \sigma_{\text{domain}} \cdot z_{\text{domain}_{i, j}}\\
\sigma_{\text{domain}} &\sim \text{HalfNormal}(0.1) \\
z_{\text{domain}_{i, j}} &\sim \text{Normal}(0, 1) \\
\text{subdomain}_{i,j,k} &= \text{domain}_{i,j} + \sigma_{\text{subdomain}} \cdot z_{\text{subdomain}_{i,j,k}} \\
\sigma_{\text{subdomain}} &\sim \text{HalfNormal}(0.1) \\
z_{\text{subdomain}_{i,j,k}} &\sim \text{Normal}(0, 1)
\end{aligned}
\label{eq:use_case2_eq}
\end{equation}
\noindent

\begin{flushleft}
\small \textit{}$\text{Correct}_{i,j,k}$ denotes the number of correct answers for the $i$-th LLM, $j$-th domain, and $k$-th subdomain, distributed as the proportion of correct answers (proportion parameter $p$) among the number of trials ($n_{i,j,k}$). The GLM assumes that the individual LLM-specific parameters are drawn from an LLM distribution. The individual domain-specific parameters are drawn from a domain distribution that is parameterized by the LLM-specific effects. The individual subdomain parameters are drawn from a subdomain distribution that is parameterized by the domain-specific effects.
\end{flushleft}

\begin{figure}[htbp]
    \centering
    \includegraphics[width=.8\textwidth]{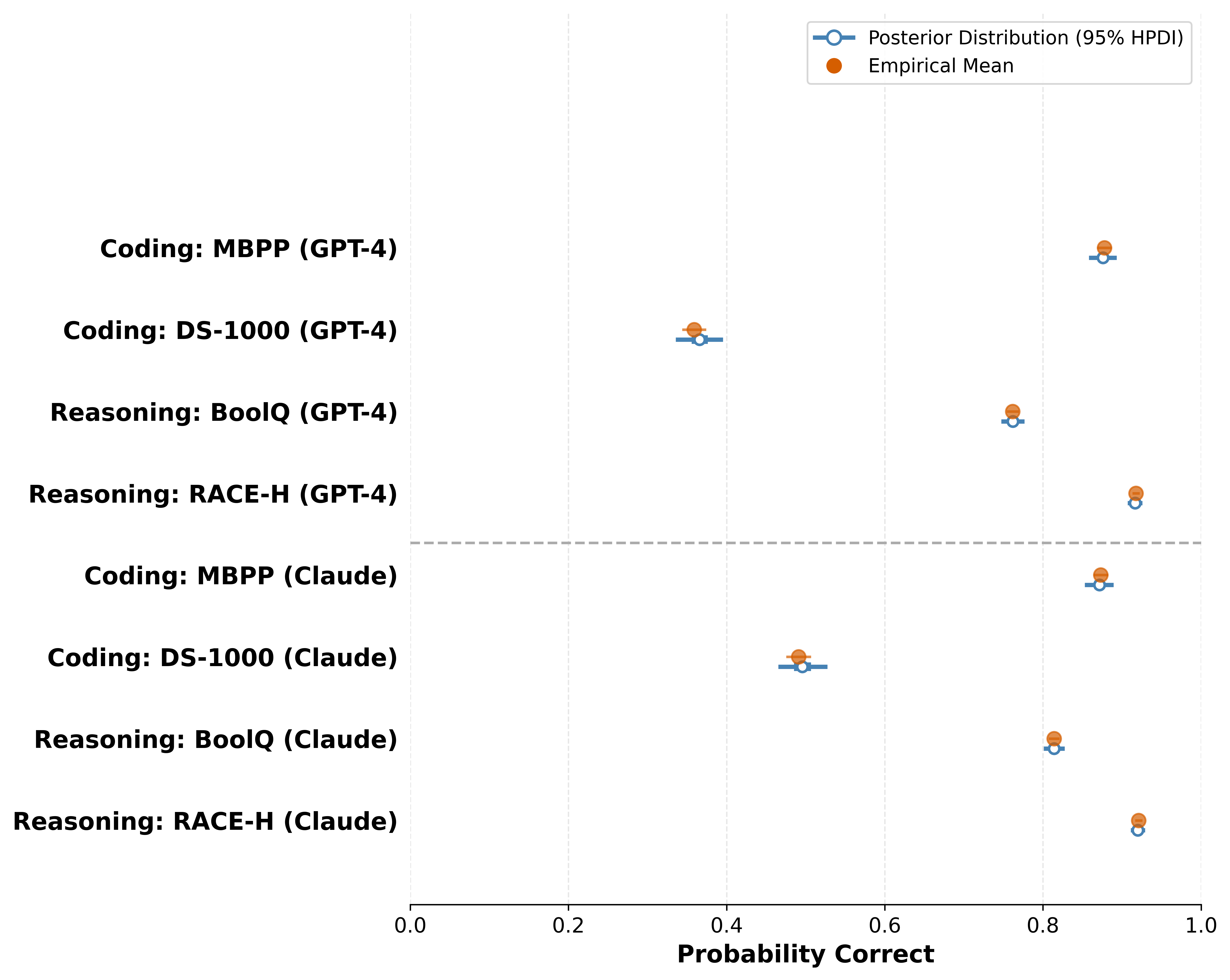}
    \caption{\textbf{Forest plot to display posterior means and 95\% HPDI (on the probability scale) for use case 2.} Focusing on model (LLM)-, domain- and subdomain-specific effects, this plot indicates the uncertainty (vertical lines, 95\% HPDI), around the posterior mean (dots) for the respective parameter estimates. Additionally, for comparison between \texttt{HiBayES} and conventional evaluation statistics approaches, we present empirical means and SEM per LLM, domain, subdomain (orange).}
    \label{fig:use_case2_forestplot}
\end{figure}

\Cref{fig:use_case2_forestplot} shows the posterior parameter estimates and compares estimated model (LLM), domain and subdomain effects and the uncertainty of the estimates (HPDI) to empirical means (arithmetic means) and standard errors of the mean (SEM) for each LLM, domain and subdomain separately (\Cref{fig:use_case2_forestplot}). We observe that the empirical SEMs are consistently narrower than the GLM estimated HPDIs. Furthermore, the empirical means and GLM estimated parameter means are similar, but not identical. The latter phenomenon, called “shrinkage,” refers to partial pooling of parameters across nested data levels in hierarchical GLMs and a resulting contraction of lower-level parameter estimates (e.g., subdomain means) towards higher-level estimates (e.g., domain means). This form of regularization imposed by the hierarchical data structure helps to produce less over-fitting and more stable estimates. In other words, the parameter estimates derived from hierarchical GLMs allow us to more confidently generalize the capability in question to new contexts beyond the specific evaluation – e.g., real-world deployment contexts.

The HPDIs for the subdomains estimated in a hierarchical GLM suggest similar differences found using the conventional approach (95\%-HPDI DS-1000 = $[.34; .40]$ vs $[.47; .53]$ and 95\%-HPDI BoolQ = $[.75; .78]$ vs $[.80; .83]$). Additionally, HPDIs allow to quantify the magnitude and uncertainty around these differences in a more nuanced and continuous way -- supporting an inference that the performance difference observed for BoolQ is smaller than for DS-1000. The conventional approach only allows us to make the discrete inference that LLM performance levels differ significantly on both subdomains. 

An additional advantage of \texttt{HiBayES} is that GLMs like the one shown in \Cref{eq:use_case2_eq} straightforwardly extend to evaluation scenarios that cover more than two domains or more than two LLMs (e.g., comparing 3 LLMs) without adjusting the statistical inference process, correcting for multiple comparisons, or using another family of tests. Since all estimated parameters are indexed, it is possible to add another number/index and estimate 3 (x3) LLM/domain parameters instead of 2 (x2) -- we illustrate this below in use case 3.

This straightforward generalization to more complex scenarios and setups, e.g., testing more than two LLMs or different levels of the data at the same time (which might often be desired), cannot be implemented within conventional statistical approaches (e.g., using \textit{t}-tests) without resorting to a different family of tests. This is because a \textit{t}-test can at maximum test differences between two samples or two repeated measures. Any additional samples or repeated measurements require an adjustment for multiple comparisons (using Family-wise Error or False-Discovery Rate Correction) – since the assumed Type I and II Error rates cannot be maintained reliably – or necessitate resorting to an entirely different family of tests (\textit{F}-tests/ANOVAs).

\subsection{Use case 3: Elicitation experiments with several LLMs evaluated on an agentic benchmark, with varying reasoning parameters, tasks at varying levels of difficulty, in low data regime}

Interest in how AI models behave in real-world, agentic settings has surged in recent years. To test these capabilities, more complex, challenging and extended benchmarks than QA benchmarks have been developed \cite{Mialon2023,Rein2025}. Of particular interest in such evaluation scenarios is to identify setups (agent scaffolds \cite{Yao2022}) that increase the success probability of models, and to make robust predictions for future behavior in deployment contexts. This extends to, for example, which tools LLMs have at their disposal, or under which level of reasoning effort they operate. The assessment of these effects has far-reaching financial implications, as token costs for advanced reasoning LLMs easily reach hundreds of US Dollars per task.

In this illustrative example, we compare the performance of several LLMs (o1, o3-mini, Claude-3.7-Sonnet and GPT-4.5-preview) on an agentic evaluation (GAIA\cite{Mialon2023}). This benchmark features 165 unique tasks at varying levels of difficulty (1: easy, 2: mid, 3: hard). We tested o1 and o3-mini at different levels of reasoning effort (1: no reasoning, 2: low, 3: intermediate or 4: high), Claude-3.7-Sonnet and GPT-4.5-preview were only evaluated in no reasoning setting. Due to token costs stemming from higher levels of reasoning efforts, this evaluation results in a low data regime at the task level (10 repeats per task and model). The dataset amounts to a total of 11,650 observations, varying substantially across the different LLMs under consideration (o1: 4,632, o3-mini: 4,664, Claude-3.7-Sonnet: 1,565, GPT-4.5-preview: 789 observations).

\begin{figure}[htbp]
    \centering
    \includegraphics[width=.7\textwidth]{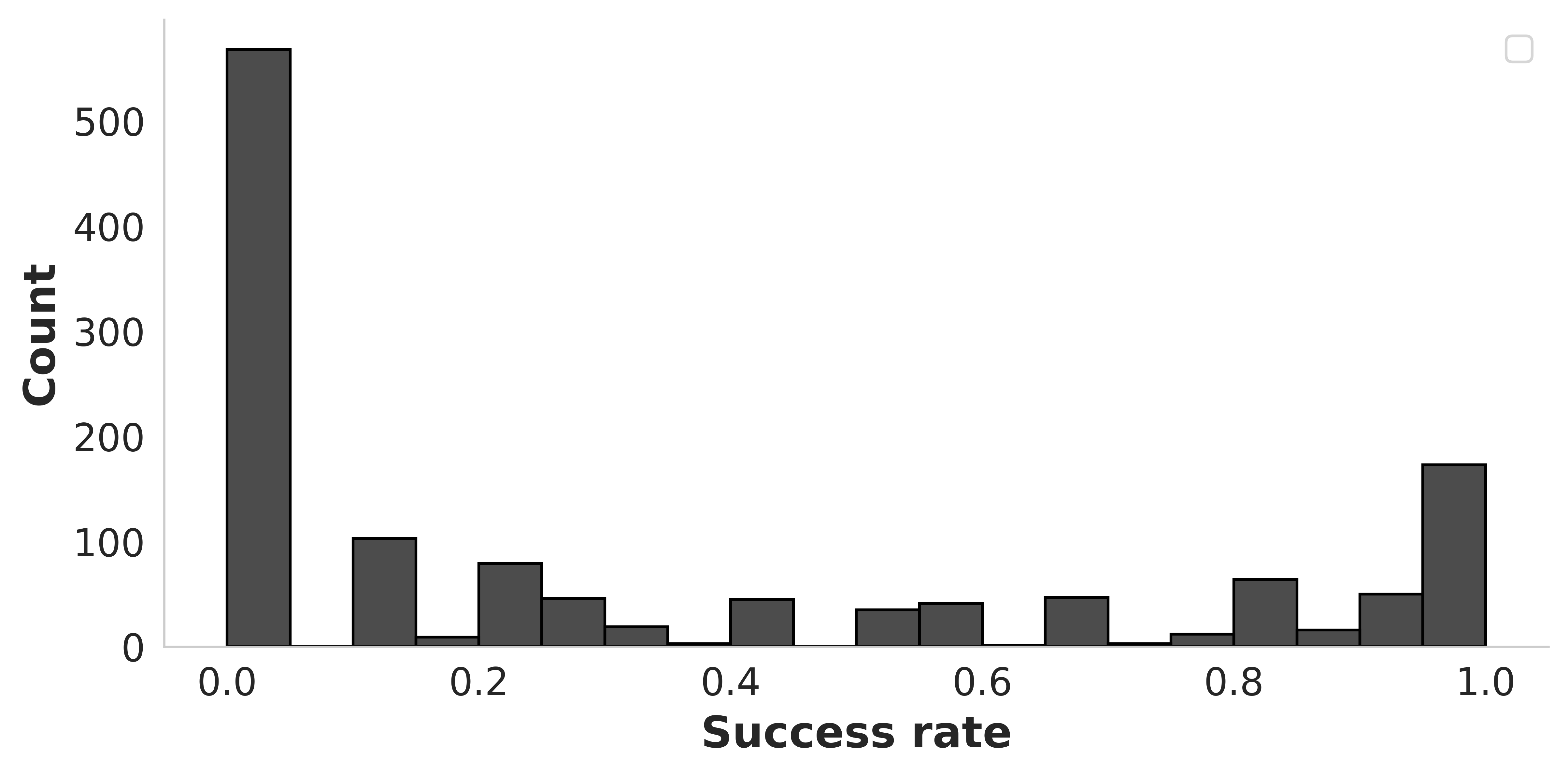}
    \caption{\textbf{Success rate distribution in GAIA.} The average success rates across repeats on different tasks (across all models, difficulty levels and reasoning efforts). GAIA success rate data is asymmetric and bimodal.}
    \label{fig:raw_data_use_case3}
\end{figure}

Investigating the raw data per task, we observed an asymmetric bimodal success rate distribution across tasks (see \Cref{fig:raw_data_use_case3}). This is a common finding in agentic evaluation settings – LLMs either pass or fail at all their attempts (success rate either 0 or 1) for most tasks, an intermediate level of success is rare. This suggests that such evaluation items are either saturated, or that they are too difficult, and therefore do not provide fine-grained diagnostic value to meaningfully differentiate LLM capabilities. Binomial GLMs struggle with over-dispersed bimodal outcome distributions (high prevalence of 0s and 1s), as success rates are assumed to be drawn from unimodal distributions. To account for such bimodal outcome data distributions, we can instead use Beta-Binomial GLMs, which assume that each task has its own task-specific success rate, drawn from a distribution shared across tasks.

The purpose of this experiment is to compare the influence of the reasoning effort parameter on LLM performance at various levels of task difficulty. We use GLM comparison 1) to show that a Beta-Binomial GLM outperforms a Binomial GLM on evaluation data with an asymmetric bimodal success rate distribution and 2) to formally provide evidence against a null hypothesis that reasoning effort does not affect model performance.

First, to estimate model performance across the reasoning effort parameter and task difficulty levels, we specified a hierarchical Beta-Binomial GLM with partial pooling, estimating a success probability for each task separately (\Cref{eq:use_case_3_betabinomial}). While the results of this GLM give an indication for the respective improvements of LLM performance related to increased reasoning efforts, the aim of our statistical modeling approach is to decide between a conventional Binomial GLM that does not estimate a success probability for each task separately, but is otherwise structurally identical to the model above (\Cref{eq:use_case_3_binomial}).

\begin{figure}[htbp]
\begin{minipage}[t]{0.48\textwidth}
\begin{equation}
\begin{aligned}
\text{Correct}_{i,j,k,d} &\sim \text{Binomial}(n_{i,j,k,d}, \theta_{i,j,k,d}) \\
\theta_{i,j,k,d} &\sim \text{Beta}(\alpha_{i,j,k,d}, \beta_{i,j,k,d}) \\
\text{logit}(p_{i,j,k,d}) &= \text{mod}_{\text{model}_i} + \text{diff}_{\text{difficulty}_d} \\ 
                          &\quad + \gamma_{\text{reasoning}_{i,d}} \cdot \text{reasoning}_{i,k,d} + \eta_{\text{task}_j} \\
\alpha_{i,j,k,d} &= p_{i,j,k,d} \cdot \varphi \\
\beta_{i,j,k,d} &= (1 - p_{i,j,k,d}) \cdot \varphi \\
\varphi &\sim \text{Gamma}(1, 0.1) \\
\bar{\mu} &= \mu_{\text{overall}} + \sigma_{\text{overall}} \cdot z_{\text{overall}} \\
\mu_{\text{overall}} &\sim \text{Normal}(0, 1) \\
\sigma_{\text{overall}} &\sim \text{HalfNormal}(1) \\
z_{\text{overall}} &\sim \text{Normal}(0, 1) \\
\text{mod}_{\text{model}} &= \mu_{\text{model, overall}} + \sigma_{\text{model}} \cdot z_{\text{model}_i} \\
\mu_{\text{model, overall}} &\sim \text{Normal}(\bar{\mu}, 1) \\
\sigma_{\text{model}} &\sim \text{HalfNormal}(0.5) \\
z_{\text{model},i} &\sim \text{Normal}(0, 1) \\
\text{diff}_{\text{difficulty}} &= \mu_{\text{difficulty, overall}} + \sigma_{\text{difficulty}} \cdot z_{\text{difficulty}_d} \\
\mu_{\text{difficulty, overall}} &\sim \text{Normal}(0, 1) \\
\sigma_{\text{difficulty}} &\sim \text{HalfNormal}(0.1) \\
z_{\text{difficulty}_d} &\sim \text{Normal}(0, 1) \\
\gamma_{\text{reasoning}} &= \mu_{\text{reasoning}_{i,d}} + \sigma_{\text{reasoning}} \cdot z_{\text{reasoning}_{i,d}} \\
\mu_{\text{reasoning}_{i,d}} &\sim \text{Normal}(0, 1) \\
\sigma_{\text{reasoning}} &\sim \text{HalfNormal}(0.1) \\
z_{\text{reasoning}_{i,d}} &\sim \text{Normal}(0, 1) \\
\eta_{\text{task}} &= \mu_{\text{task, overall}} + \sigma_{\text{task}} \cdot z_{\text{task}_j} \\
\sigma_{\text{task}} &\sim \text{HalfNormal}(0.5) \\
z_{\text{task}_j} &\sim \text{Normal}(0, 1).
\end{aligned}
\label{eq:use_case_3_betabinomial}
\end{equation}
\begin{flushleft}
\small \textit{}$\text{Correct}_{i,j,k,d}$ denotes the number of correct answers for the $i$-th LLM, $j$-th task, $k$-th reasoning level, and $d$-th difficulty level, distributed as the proportion of correct answers separately per task (parameter $\theta$ drawn from a Beta distribution) among the number of trials ($n_{i,j,k,d}$).
\end{flushleft}
\end{minipage}
\hfill 
\begin{minipage}[t]{0.45\textwidth}
\begin{equation}
\begin{aligned}
\text{Correct}_{i,j,k,d} &\sim \text{Binomial}(n_{i,j,k,d}, p_{i,j,k,d}) \\
\text{logit}(p_{i,j,k,d}) &= \text{mod}_{\text{model}_i} + \text{diff}_{\text{difficulty}_d} \\ 
                         &\quad + \gamma_{\text{reasoning}_{i,d}} \cdot \text{reasoning}_{i,k,d} + \eta_{\text{task}_j} \\
\bar{\mu} &= \mu_{\text{overall}} + \sigma_{\text{overall}} \cdot z_{\text{overall}} \\
\mu_{\text{overall}} &\sim \text{Normal}(0, 1) \\
\sigma_{\text{overall}} &\sim \text{HalfNormal}(1) \\
z_{\text{overall}} &\sim \text{Normal}(0, 1) \\
\text{mod}_{\text{model}} &= \mu_{\text{model, overall}} + \sigma_{\text{model}} \cdot z_{\text{model}_i} \\
\mu_{\text{model, overall}} &\sim \text{Normal}(\bar{\mu}, 1) \\
\sigma_{\text{model}} &\sim \text{HalfNormal}(0.5) \\
z_{\text{model},i} &\sim \text{Normal}(0, 1) \\
\text{diff}_{\text{difficulty}} &= \mu_{\text{difficulty, overall}} + \sigma_{\text{difficulty}} \cdot z_{\text{difficulty}_d} \\
\mu_{\text{difficulty, overall}} &\sim \text{Normal}(0, 1) \\
\sigma_{\text{difficulty}} &\sim \text{HalfNormal}(0.1) \\
z_{\text{difficulty},d} &\sim \text{Normal}(0, 1) \\
\gamma_{\text{reasoning}} &= \mu_{\text{reasoning}_{i,d}} + \sigma_{\text{reasoning}} \cdot z_{\text{reasoning}_{i,d}} \\
\mu_{\text{reasoning}_{i,d}} &\sim \text{Normal}(0, 1) \\
\sigma_{\text{reasoning}} &\sim \text{HalfNormal}(0.1) \\
z_{\text{reasoning}_{i,d}} &\sim \text{Normal}(0, 1) \\
\eta_{\text{task}} &= \mu_{\text{task, overall}} + \sigma_{\text{task}} \cdot z_{\text{task}_j} \\
\sigma_{\text{task}} &\sim \text{HalfNormal}(0.5) \\
z_{\text{task}_j} &\sim \text{Normal}(0, 1)
\end{aligned}
\label{eq:use_case_3_binomial}
\end{equation}
\begin{flushleft}
\small \textit{}$\text{Correct}_{i,j,k,d}$ denotes the number of correct answers for the $i$-th LLM, $j$-th task, $k$-th reasoning level, and $d$-th difficulty level, distributed as the proportion of correct answers (proportion parameter $p$) among the number of trials ($n_{i,j,k,d}$).
\end{flushleft}
\end{minipage}
\end{figure}

Second, we want to know whether the Reasoning Beta-Binomial GLM is a good explanation of the data, relative to alternative GLMs. Such inference about the underlying data generating process is not possible solely based on investigating the parameter estimates. To evaluate the question of whether reasoning effort improves the success rate of LLMs in GAIA, we need to employ a GLM comparison to formally test the null hypothesis that reasoning effort does not affect LLM performance. This null hypothesis is incorporated by a hierarchical Binomial GLM with partial pooling that does not feature information on reasoning efforts and task difficulty levels (\Cref{eq:use_case_3_null}). 

\begin{figure}[htbp]
    \centering
    \includegraphics[width=.7\textwidth]{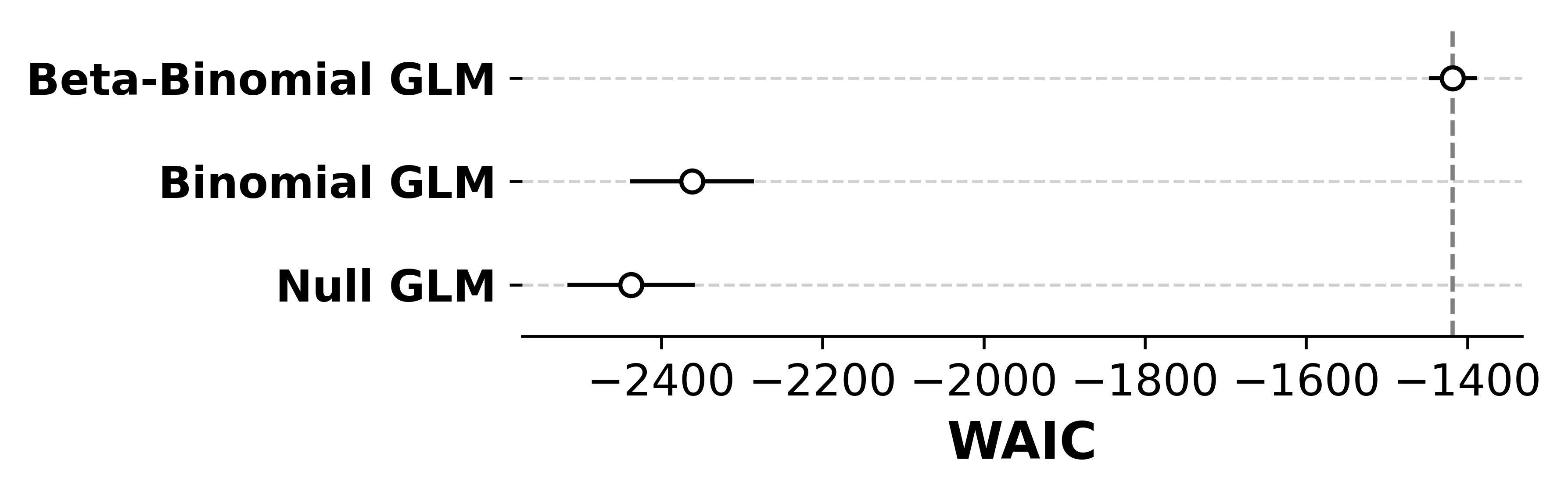}
    \caption{\textbf{GLM comparison using WAIC for use case 3.} The plot shows the WAIC values for the three GLMs. Closer to 0 (higher) WAIC values indicate better model fit. The Reasoning Beta-Binomial GLM, incorporating model and difficulty level specific effects of reasoning efforts, provides a better account for the data than the Reasoning Binomial GLM (not accounting for task level variation in success probability) and the Null GLM (not incorporating model and difficulty level specific effects of reasoning efforts).
}
    \label{fig:use_case3_model_comp}
\end{figure}

After estimating the posterior distribution for the Null GLM, we compare the predictive accuracy of all three GLMs (\Cref{eq:use_case_3_betabinomial} - \Cref{eq:use_case_3_null}) using WAIC. In general, a WAIC closer to 0 indicates better fit and better predictive accuracy (better explanation of the data than alternative GLMs). The GLM comparison based on WAIC indicates that the Reasoning Beta-Binomial GLM accounts for the data better than the Reasoning Binomial GLM and the Null GLM (\Cref{fig:use_case3_model_comp}). This finding provides evidence that reasoning effort is a relevant aspect of the agentic setup.

\begin{equation}
\begin{aligned}
\text{Correct}_{i,j,d} &\sim \text{Binomial}(n_{i,j,d}, p_{i,j,d}) \\
\text{logit}(p_{i,j,d}) &= \text{mod}_{\text{model}_i} + \text{diff}_{\text{difficulty}_d} + \eta_{\text{task}_j} \\
\bar{\mu} &= \mu_{\text{overall}} + \sigma_{\text{overall}} \cdot z_{\text{overall}} \\
\mu_{\text{overall}} &\sim \text{Normal}(0, 1) \\
\sigma_{\text{overall}} &\sim \text{HalfNormal}(1) \\
z_{\text{overall}} &\sim \text{Normal}(0, 1) \\
\text{mod}_{\text{model}} &= \mu_{\text{model, overall}} + \sigma_{\text{model}} \cdot z_{\text{model}_i} \\
\mu_{\text{model, overall}} &\sim \text{Normal}(\bar{\mu}, 1) \\
\sigma_{\text{model}} &\sim \text{HalfNormal}(1) \\
z_{\text{model}_i} &\sim \text{Normal}(0, 1) \\
\text{diff}_{\text{difficulty}} &= \mu_{\text{difficulty, overall}} + \sigma_{\text{difficulty}} \cdot z_{\text{difficulty}_d} \\
\mu_{\text{difficulty, overall}} &\sim \text{Normal}(0, 1) \\
\sigma_{\text{difficulty}} &\sim \text{HalfNormal}(0.1) \\
z_{\text{difficulty}_d} &\sim \text{Normal}(0, 1) \\
\eta_{\text{task}} &= \mu_{\text{task, overall}} + \sigma_{\text{task}} \cdot z_{\text{task}_j} \\
\sigma_{\text{task}} &\sim \text{HalfNormal}(0.5) \\
z_{\text{task}_j} &\sim \text{Normal}(0, 1)
\end{aligned}
\label{eq:use_case_3_null}
\end{equation}

\begin{flushleft}
\small \textit{}$\text{Correct}_{i,j,d}$ denotes the number of correct answers for the $i$-th LLM, $j$-th task, and $d$-th difficulty level, distributed as the proportion of correct answers (proportion parameter $p$) among the number of trials ($n_{i,j,d}$).
\end{flushleft}

\begin{figure}[htbp]
    \centering
    \includegraphics[width=.7\textwidth]{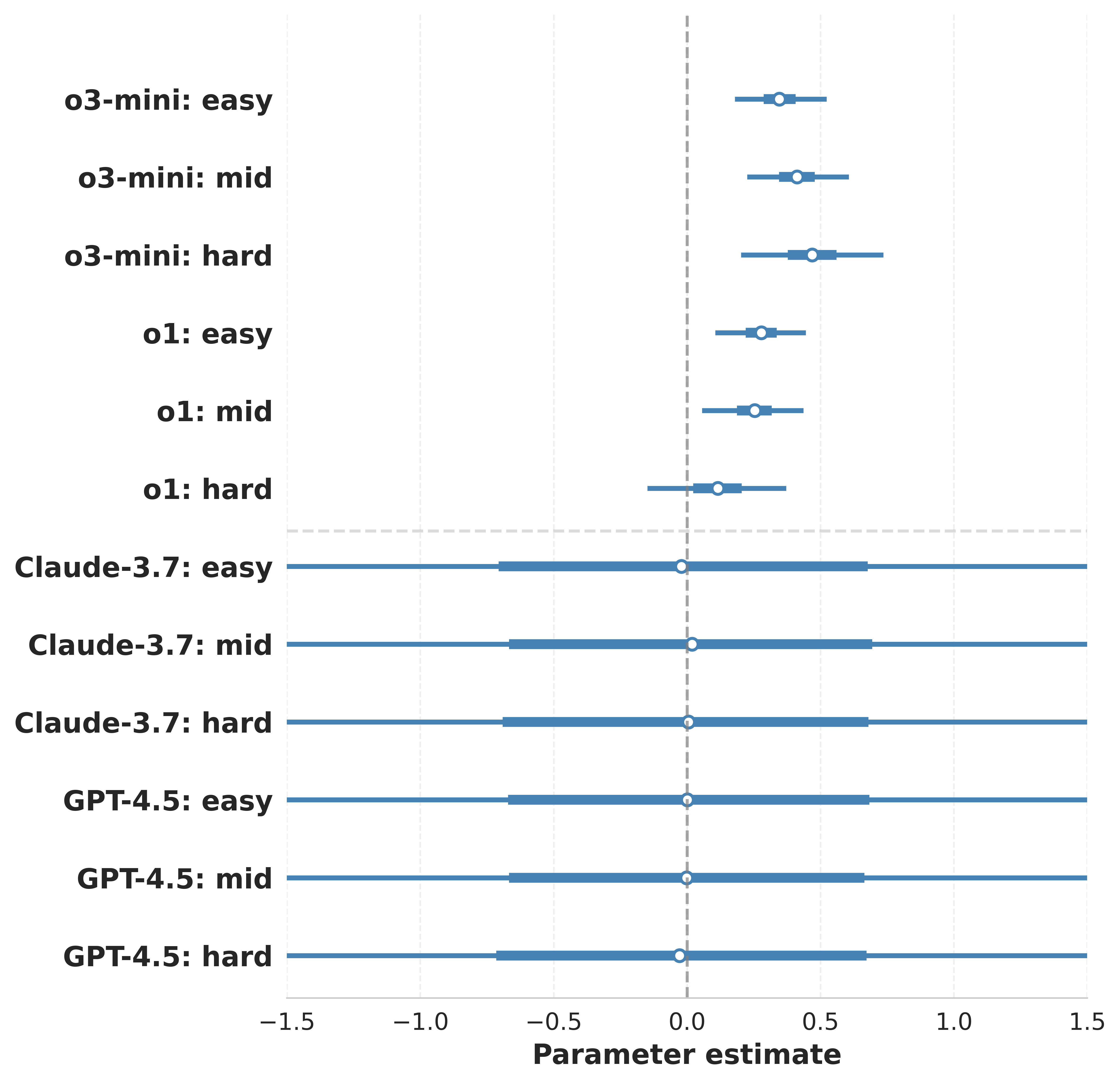}
    \caption{\textbf{Forest plot for the effect of reasoning effort (none, low, intermediate, high) and task difficulty (easy, mid, hard) on performance of 4 different LLMs (use case 3).} The plot displays posterior means and 95\% HPDI (vertical lines) for the respective parameter estimates (on logit scale).
}
    \label{fig:use_case3_forest_plot}
\end{figure}

We next want to understand more about the specific effect of reasoning efforts and task difficulty on LLM performance. To this end, we extract parameter estimates of the Beta-Binomial GLM defined in \Cref{eq:use_case_3_betabinomial}. The forest plot (\Cref{fig:use_case3_forest_plot}) displays the posterior distributions for estimated linear effects of reasoning effort ($\gamma_{\text{reasoning}}$) for each LLM at different levels of task difficulty and provides the uncertainty of these estimates. While the parameter estimates for o3-mini and o1 (except in hard tasks) are positive and non-zero (HPDI does not contain 0) across all levels of task difficulty, o3-mini seems to benefit slightly more from reasoning across task difficulty levels. For the two LLMs tested without variation of the reasoning effort (Claude-3.7-Sonnet and GPT-4.5-preview), these effects are not different from 0 (consistent with the notion that these LLMs were only tested under non-reasoning settings).

In combination, the GLM comparison and the parameter estimates indicate that reasoning improves performance across task difficulty. The findings suggest that high reasoning effort provide performance gains in o3-mini, especially on hard tasks -- but the costs of high reasoning may be less justified for o1, where reasoning effort was equally effective on easy and mid-level tasks, but not effective on hard tasks.

We also present formal evidence that Beta-Binomial GLMs provide substantially better fit to agentic evaluation data that are characterized by over-dispersion of 0s and 1s (i.e., models either failing or passing at all attempts).

\section{Conclusion}

This paper introduced \texttt{HiBayES}, a robust and flexible statistical modeling framework for conducting statistical analysis on evaluation results. \texttt{HiBayES} is centered around multilevel (hierarchical) Bayesian modeling using Generalized Linear Models, optimal uncertainty quantification across nested hierarchical levels of the data, and formal model comparison to infer AI system capabilities. Our approach overcomes fundamental limitations of common practices and conventional approaches to evaluation statistics, such as "highest number is best", based on point estimates and lacking uncertainty quantification, or "flat" statistical modeling (e.g., $t$-tests) using pre-averaged data, without conducting formal model comparisons. These practices affect researchers' ability to make correct predictions about future model behavior in production and lead to an overconfident interpretation of the numerical performance differences between models.

\texttt{HiBayES} can reliably be used across evaluations with varying data quantities and delivers robust estimates even in low-data regimes (fewer than 20 data points per evaluation). We anticipate that such scenarios will become increasingly prevalent when evaluating new agentic or reasoning model capabilities, where generating datasets can easily consume tokens worth hundreds of US Dollars per task. \texttt{HiBayES} therefore represents an efficient, economic alternative to approaches that rely on obtaining and pre-averaging large quantities of evaluation data in order to produce stable estimates. We present formal evidence that Beta-Binomial likelihood functions provide substantially better fit in agentic evaluation settings, which are often characterized by over-dispersion of 0s and 1s (models either failing or passing at all attempts). We therefore recommend using Beta-Binomial GLMs for agentic evaluation data.

\texttt{HiBayES} is a principled, highly flexible, and generalizable statistical framework that allows researchers and practitioners to robustly test and adapt to a multitude of evaluation scenarios with statistical rigor – e.g., comparing the performance of two or more models on a number of different domains, subdomains, or repeats. \texttt{HiBayES} can adequately capture correlation structures and mirror the hierarchical context in which evaluation data is acquired, enabling more precise estimation of model capabilities and prediction of future model behavior. Our hope is that the \texttt{HiBayES} framework will be adopted widely by the AI and evaluation research community, that it will have a positive impact on the statistical rigor and the quality of statistical inferences drawn in AI evaluations, and that it may help reduce the amount of evaluation data needed while still enabling valid conclusions about model capabilities.

Potential extensions of the framework relate to creating adaptive evaluation systems that store and update posterior beliefs on general model capabilities in real time as models are evaluated on new benchmarks (serving as priors for new evaluations) – either privately within companies or openly accessible to other researchers. This principled scientific approach would enhance precision and may help reduce uncertainty over model capability estimates across time and with growing experience. Another promising avenue for multilevel Bayesian modeling in evaluation statistics is predictive evaluations: Improving methods for estimating correlations between related tasks and domains – for instance, is performance in the MMLU \cite{Hendrycks2020} subdomain \textit{college computer science} more strongly associated with a model’s InterCode CTF \cite{Yang2023} capability than MMLU \cite{Hendrycks2020} \textit{philosophy} subdomain knowledge? Other predictive evaluation questions relate to generalizing results of automated evaluation tasks to red-teaming exercises or from automated evaluations to human uplift results, or using one model’s capabilities to predict expected capabilities of another model from the same developer or family, before the second model is tested.

Such an approach would allow leveraging hierarchical evaluation data structures and data clusters to enhance predictive accuracy and generalization of evaluation statistics. Systematically decomposing evaluations by identifying their constituting parts and the associative structure relating abilities needed to solve one evaluation and to succeed at other evaluation domains and tasks may provide deep insights into the latent structure of model capabilities and may facilitate designing more efficient and adaptive evaluations.

\newpage

\section{Appendix}\label{sec:appendix}

Define and fit \texttt{HiBayES} multilevel Bayesian GLMs in 8 steps. More detailed information can be found in \cite{Gelman2013} and \cite{McElreath2020a,McElreath2020b}.

\begin{figure}[htbp]
    \centering
    \adjustbox{trim={0 3cm 2.75cm 0.5cm}, clip}{\includegraphics[width=1\textwidth]{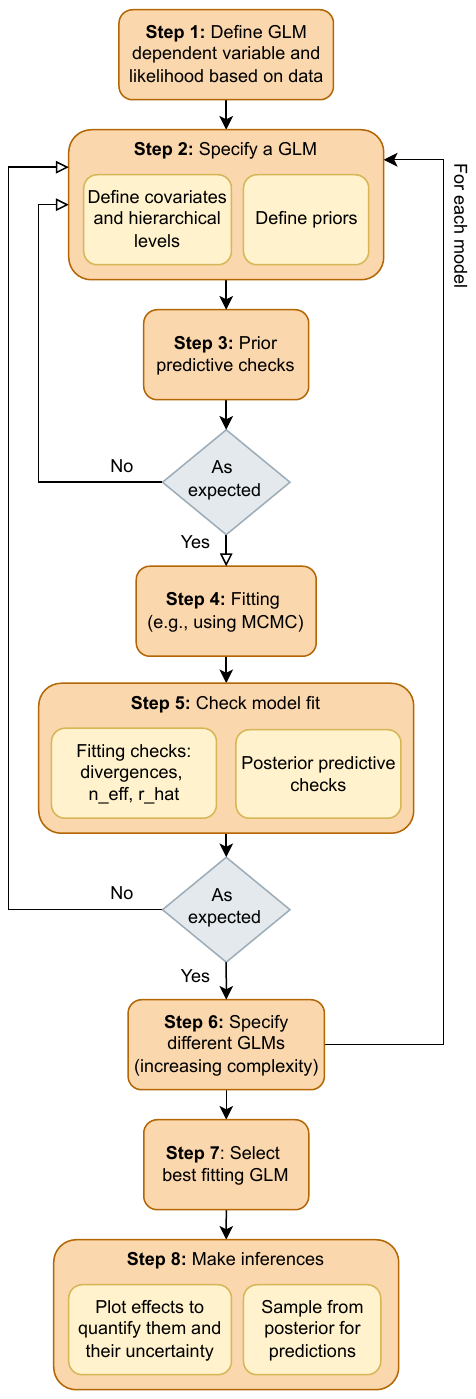}}
    \label{fig:steps_for_GLM}
\end{figure}

\newpage

\section*{Acknowledgments}
We thank Jake Pencharz, Hannah Kirk and Geoffrey Irving for helpful comments and discussions on earlier versions of the manuscript. 

\newpage

\printbibliography

\end{document}